\documentclass[11pt,a4paper]{article}
\usepackage{authblk}
\usepackage[hyphens,spaces,obeyspaces]{url}

\usepackage[hyperref]{emnlp2018}
\usepackage{times}
\usepackage{latexsym}
\usepackage{url}
\usepackage{booktabs}
\usepackage{amsmath}
\usepackage{multirow}
\usepackage{url}
\usepackage{float}
\usepackage{graphicx}
\usepackage{tikz}
\usepackage{amssymb}

\usepackage[normalem]{ulem}

\aclfinalcopy 

\title{On Incorporating Structural Information to improve Dialogue Response
Generation}

\author[1]{\textbf{Nikita Moghe}\thanks{The work was done by Nikita and Priyesh  at Indian Institue of Technology Madras. Email \texttt{nikita.moghe@ed.ac.uk}}}
\author[2]{\textbf{Priyesh Vijayan}}
\author[3,4]{\textbf{Balaraman Ravindran}}
\author[3,4] {\textbf{Mitesh M. Khapra}}
\affil[1]{School of Informatics, University of Edinburgh}
\affil[2]{School of Computer Science, McGill University and Mila}
\affil[3]{Indian Institute of Technology Madras}
\affil[4]{Robert Bosch Centre for Data Science and Artificial Intelligence (RBC-DSAI), \protect\\ Indian Institute of Technology Madras}

\date{}

\begin{document}
\maketitle
\begin{abstract}
We consider the task of generating dialogue responses from background knowledge comprising of domain specific resources. Specifically, given a conversation around a movie, the task is to generate the next response based on background knowledge about the movie such as the plot, review, Reddit comments \textit{etc}. This requires capturing structural, sequential and semantic information from the conversation context and the background resources. This is a new task and has not received much attention from the community. We propose a new architecture that uses the ability of BERT to capture deep contextualized representations in conjunction with explicit structure and sequence information. More specifically, we use (i) Graph Convolutional Networks (GCNs) to capture structural information, (ii) LSTMs to capture sequential information and (iii) BERT for the deep contextualized representations that capture semantic information. We analyze the proposed architecture extensively. To this end, we propose a plug-and-play \textit{Semantics-Sequences-Structures} (\textit{SSS}) framework which allows us to effectively combine such linguistic information. Through a series of experiments we make some interesting observations. First, we observe that the popular adaptation of the GCN model for NLP tasks where structural information (GCNs) was added on top of sequential information (LSTMs) performs poorly on our task. This leads us to explore interesting ways of combining semantic and structural information to improve the performance. 
Second, we observe that while BERT already outperforms other deep contextualized representations such as ELMo, it still benefits from the additional structural information explicitly added using GCNs. This is a bit surprising given the recent claims that BERT already captures structural information. Lastly, the proposed \textit{SSS} framework gives an improvement of 7.95\% over the baseline.
\end{abstract}

\section{Introduction}
Neural conversation systems which treat dialogue response generation as a sequence generation task \cite{neural_conversation_model} often produce generic and incoherent responses \cite{shao-etal-2017-generating}. The primary reason for this is that, unlike humans, such systems do not have any access to background knowledge about the topic of conversation. For example, while chatting about movies, we use our background knowledge about the movie in the form of plot details, reviews and comments that we might have read. To enrich such neural conversation systems, some recent works \cite{moghe-etal-2018-towards,wizard_of_wikipedia,zhou-etal-2018-dataset} incorporate external knowledge in the form of documents which are relevant to the current conversation. For example, \cite{moghe-etal-2018-towards}, released a dataset containing conversations about movies where every alternate utterance is extracted from a background document about the movie. This background document contains plot details, reviews and Reddit comments about the movie. The focus thus shifts from sequence generation to identifying relevant snippets from the background document and modifying them suitably to form an appropriate response given the current conversational context. 

\begin{figure}
\centering
\begin{tikzpicture}
\node[draw, text width=7.3cm] at (-6,0) {\textbf{Source Doc:} ... At this point James Braddock (Russel Crowe) was a light heavyweight boxer, who was forced to retired from the ring after breaking his hand in his last fight. \textbf{His wife Mae had prayed for years that he would quit boxing, before becoming permanently injured}. ...

\textbf{Conversation:} \\
Speaker 1(N): Yes very true, this is a real rags to riches story. Russell Crowe was excellent as usual. \\
Speaker 2(R): Russell Crowe owns the character of James Bradock, the unlikely hero who makes the most of his second chance. He's a good fighter turned hack. \\

Speaker 1(N): Totally! Oh by the way do you remember his wife ... how she wished he would stop\\

Speaker 2(P): His wife Mae had prayed for years that he would quit boxing, before becoming permanently injured.\\
};
\node[draw] at (-6,-6.2){\includegraphics[scale=0.37]{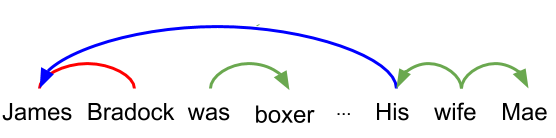}};
\end{tikzpicture}
\caption{Sample conversation from the Holl-E Dataset.  For simplicity, we show only a few of the edges. The edge in blue corresponds to co-reference edge, the edges in green are dependency edges and the edge in red is the entity edge. }
\label{dataset_example}
\end{figure}

Intuitively, any model for this task should exploit semantic, structural and sequential information from the conversation context and the background document. For illustration, consider the chat shown in Figure \ref{dataset_example} from the Holl-E movie conversations dataset \cite{moghe-etal-2018-towards}. In this example, Speaker 1 nudges Speaker 2 to talk about how James's wife was irritated because of his career. The right response to this conversation comes from the line beginning at \textit{``His wife Mae \dots''}. However, to generate this response, it is essential to understand that (i) \textit{His} refers to James from the previous sentence; (ii) \textit{quit boxing} is a contiguous phrase, and (iii) \textit{quit} and \textit{he would stop} mean the same. We need to exploit (i) \textbf{structural} information, such as,  the co-reference edge between \textit{His-James} (ii) the \textbf{sequential} information in \textit{quit boxing} and (iii) the \textbf{semantic} similarity (or synonymy relation) between \textit{quit} and \textit{he would stop}.


To capture such multi-faceted information from the document and the conversation context we propose a new architecture which combines BERT with explicit sequence and structure information. We start with the deep contextualized word representations learnt by BERT which capture distributional semantics. We then enrich these representations with sequential information by allowing the words to interact with each other by passing them through a bidirectional LSTM as is the standard practice in many NLP tasks. Lastly, we add explicit structural information in the form of dependency graphs, co-reference graphs, and entity co-occurrence graphs. To allow interactions between words related through such structures, we use GCNs which essentially aggregate information from the neighborhood of a word in the graph. 

Of course, combining BERT with LSTMs in itself is not new and has been tried in the original work \cite{devlin-etal-2019-bert} for the task of Named Entity Recognition. Similarly, the work in \cite{bastings-etal-2017-graph} combines LSTMs with GCNs for the task of machine translation. To the best of our knowledge this is the first work which combines BERT with explicit structural information. We investigate several interesting questions in the context of dialogue response generation. For example, (i) Are BERT-based models best suited for this task? (ii) Should BERT representations be enriched with sequential information first or structural information? (iii) Are dependency graph structures more important for this task or entity co-occurence graphs? (iv) Given the recent claims that BERT captures syntactic information, does it help to explicitly enrich it with syntactic information using GCNs?

To systematically investigate such questions we propose a simple plug-and-play \textit{Semantics-Sequences-Structures} (\textit{SSS}) framework which allows us to combine different semantic representations (GloVe, BERT, ELMo) with different structural priors (dependency graphs, co-reference graphs, \textit{etc.}). It also allows us to use different ways of combining structural and sequential information, \textit{e.g.}, LSTM first followed by GCN or vice versa or both in parallel. Using this framework we perform a series of experiments on the \textit{Holl-E} dataset and make some interesting observations. First, we observe that the conventional adaptation of GCNs for NLP tasks, where contextualized embeddings obtained through LSTMs are fed as input to a GCN, exhibits poor performance. To overcome this, we propose some simple alternatives and show that they lead to better performance. Second, we observe that while BERT performs better than GloVe and ELMo, it still benefits from explicit structural information captured by GCNs. We find this interesting because some recent works \cite{tenney-etal-2019-bert,jawahar-etal-2019-bert,hewitt-manning-2019-structural} suggest that BERT captures syntactic information, but our results suggest that there is still more information to be captured by adding explicit structural priors. Third, we observe that certain graph structures are more useful for this task than others. Lastly, our best model which uses a specific combination of semantic, sequential and structural information improves over the baseline by 7.95\%.

\section{Related work}
There is active interest in using external knowledge to improve informativeness of responses for goal-oriented as well as chit-chat conversations \cite{lowe2015,KGNCM,moghe-etal-2018-towards,wizard_of_wikipedia}. Even the teams participating in the annual Alexa Prize competition \cite{alexa_prize} have benefited by using several knowledge resources. This external knowledge can be in the form of knowledge graphs or unstructured texts such as documents.

Many NLP systems including conversation systems use RNNs as their basic building block which typically capture $n$-gram or sequential information. Adding structural information through tree-based structures \cite{tai-etal-2015-improved} or graph based structures \cite{marcheggiani-titov-2017-encoding} on top of this has shown improved results on several tasks. For example, GCNs have been used to improve neural machine translation \cite{marcheggiani-etal-2018-exploiting} by exploiting the semantic structure of the source sentence. Similarly, GCNs have been used with dependency graphs to incorporate structural information for semantic role labelling \cite{marcheggiani-titov-2017-encoding}, neural machine translation \cite{bastings-etal-2017-graph} and entity relation information in question answering \cite{de-cao-etal-2019-question} and  temporal information for neural dating of documents \cite{vashishth-etal-2018-dating}.

There have been advances in learning deep contextualized word representations \cite{peters-etal-2018-deep,devlin-etal-2019-bert} with a hope that such representations will implicitly learn structural and relational information with interaction between words at multiple layers \cite{jawahar-etal-2019-bert,peters-etal-2018-dissecting}. These recent developments have led to many interesting questions about the best way of exploiting rich information from sentences and documents. We try to answer some of these questions in the context of background aware dialogue response generation.

\section{Background}
\label{sec:background}
In this section, we provide a background on how GCNs have been leveraged in NLP to incorporate different linguistic structures.

\label{sec:gcn-background}

The Syntactic-GCN proposed in \cite{marcheggiani-titov-2017-encoding} is a GCN \cite{kipf2016semi} variant which can model multiple edge types and edge directions. It can also dynamically determine the importance of an edge. They only work with one graph structure at a time with the most popular structure being the dependency graph of a sentence. For convenience, we refer to Syntactic-GCNs as GCNs from here on.

Let $G$ denote a graph defined on a text sequence (sentence, passage or document) with nodes as words and edges representing a directed relation between words. Let $\mathcal{N}$ denote a dictionary of list of neighbors with $\mathcal{N}(v)$ referring to the neighbors of a specific node $v$, including itself (self-loop). Let $dir(u,v) \in \{in, out, self\}$ denote the direction of the edge, $(u,v)$. Let $\mathcal{L}$ be the set of different edge types and let $L(u,v) \in \mathcal{L}$ denote the label of the edge, $(u,v)$. The $(k+1)$-hop representation of a node $v$ is computed as
\begin{equation}
h^{(k+1)}_v=\sigma(\sum_{u \in\mathcal{N}(v)}g^{(k)}_{(u,v)}(W_{dir(u,v)}^{(k)}{h}_u^{(k)}+ b_{L(u,v)}^{(k)})
\label{eqn:final-syntactic-gcn}    
\end{equation}
where $\sigma$ is the activation function, $g_{(u,v)} \in \mathbb{R}$ is the predicted importance of the edge $(u,v)$ and $h_v \in \mathcal{R}^m$ is node, $v$'s embedding. $W_{dir(u,v)} \in \{W_{in},W_{out}, W_{self}\}$ depending on the direction $dir(u,v)$ and  $W_{in}$, $W_{self}$ and $W_{out} \in \mathcal{R}^{m*m}$. The importance of an edge $g_{(u,v)}$ is determined by an edge gating mechanism w.r.t. the node of interest, $u$ as given below:
\begin{equation}
g_{(u,v)} = sigmoid \big( h_{u} \ . \ W_{dir(u,v)} + b_{L(u,v)} \big)    
\label{eqn:edge-gating}
\end{equation}
In summary, a GCN computes new representation of a node $u$ by aggregating information from it's neighborhood $\mathcal{N}(v)$. When $k$=0, the aggregation happens only from immediate neighbors, i.e., 1 hop neighbors. As the value of $k$ increases the aggregation implicitly happens from a larger neighborhood.

\section{Proposed Model}
\label{sec:proposed}
Given a document $D$ and a conversational context $Q$ the task is to generate the response $\mathbf{y} = y_1, y_2, ...., y_m$. This can be modeled as the problem of finding a $\mathbf{y}^*$ that maximizes the probability $P(\mathbf{y}$\textbar $D,Q)$ which can be further decomposed as
\begin{align*}
\mathbf{y}^*= \arg\max_\mathbf{y} \prod_{t=1}^{m} P(y_t | y_1,...,y_{t-1},Q,D)
\end{align*}
As has become a standard practice in most NLG tasks, we model the above probability using a neural network comprising of an encoder, a decoder, an attention mechanism and a copy mechanism. The copy mechanism essentially helps to directly copy words from the document $D$ instead of predicting them from the vocabulary. Our main contribution is in improving the document encoder where we use a plug-and-play framework to combine semantic, structural and sequential information from different sources. This enriched document encoder could be coupled with any existing model. In this work, we couple it with the popular GTTP model \cite{see-etal-2017-get} as used by the authors of the \textit{Holl-E} dataset. In other words, we use the same attention mechanism, decoder and copy mechanism as GTTP but augment it with an enriched document encoder. Below, we first describe the document encoder and then very briefly describe the other components of the model.  We also refer the reader to the supplementary material for more details.

\subsection{Encoder}
Our encoder contains a semantics layer, a sequential layer and a structural layer to compute a representation for the document words which is a sequence of words $w_1$, $w_2$, ..., , $w_m$. We refer to this as a plug-and-play document encoder simply because it allows us to plug in different semantic representations, different graph structures and different simple but effective mechanisms for combining structural and semantic information. 

\noindent\textbf{Semantics Layer:} 
Similar to almost all NLP models, we capture semantic information using word embeddings. In particular, we utilize the ability of BERT to capture deep contextualised representations and later combine it with explicit structural information. This allows us to evaluate (i) whether BERT is better suited for this task as compared to other embeddings such as ELMo and GloVe and (ii) whether BERT already captures syntactic information completely (as claimed by recent works) or can it benefit form additional syntactic information as described below.

\noindent\textbf{Structure Layer:} 
To capture structural information we propose multi-graph GCN, \textit{M-GCN}, a simple extension of GCN to extract relevant multi-hop multi-relational dependencies from multiple structures/graphs efficiently. In particular, we generalize $G$ to denote a labelled multi-graph, \textit{i.e.}, a graph which can contain multiple (parallel) labelled edges between the same pair of nodes. Let $\mathcal{R}$ denote the set of different graphs (structures) considered and let $G=\{\mathcal{N}_1, \mathcal{N}_2\dots \mathcal{N}_{|\mathcal{R}|}\}$ be  a set of dictionary of neighbors from the $|\mathcal{R}|$ graphs. We extend the Syntactic GCN defined in Eqn: \ref{eqn:final-syntactic-gcn} to multiple graphs by having $|\mathcal{R}|$ graph convolutions at each layer as given in Eqn: \ref{eqn:m-gcn}. Here, $g\_conv(\mathcal{N})$ is the graph convolution defined in Eqn: \ref{eqn:final-syntactic-gcn} with $\sigma$ as the identity function. Further, we remove the individual node (or word) $i$ from the neighbourhood list $\mathcal{N}(i)$ and model the node information separately using the parameter $W_{self}$.
\begin{equation}
    h^{(k+1)}_i  = ReLU \big( (h^{(k)}_iW^{(k)}_{self} + \sum_{\mathcal{N} \in G } g\_conv(\mathcal{N}) \big) 
\label{eqn:m-gcn}
\end{equation}
This formulation is advantageous over having $|\mathcal{R}|$ different GCNs as it can extract information from multi-hop pathways and can use information across different graphs with every GCN layer (hop). Note that $h^{0}_i$ is the embedding obtained for word $v$ from the semantic layer. For ease of notation, we use the following functional form to represent the final representation computed by by M-GCN after $k$-hops starting from the initial representation $h^{0}_i$, given $G$. 
\begin{equation*}
    h_i = M\textnormal{-}GCN(h_i^0, G, k)
\end{equation*}

\noindent\textbf{Sequence Layer:} 
The purpose of this layer is to capture sequential information. Once again, following standard practice, we pass the word representations computed by the previous layer through a bidirectional LSTM to compute a sequence contextualized representation for each word. As described in the next subsection, depending upon the manner in which we combine these layers, the previous layer could either be the structure layer or the semantics layer. 

\begin{figure*}
\centering
    \resizebox{\textwidth}{7.7 cm}{\includegraphics{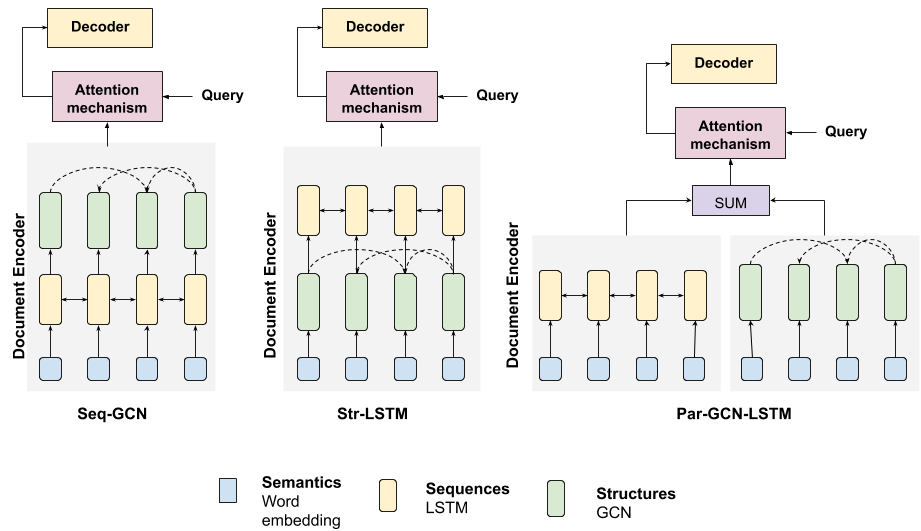}}
    \caption{The \textit{SSS} framework}
    \label{img:sss-design}
\end{figure*}

\subsection{Combining structural and sequential information}
\label{sec:SSS Framework}
As mentioned earlier, for a given document $D$ containing words $w_1, w_2, w_3, \dots, w_m$, we first obtain word representations $x_1, x_2, x_3, \dots, x_m$ using BERT (or ELMo or GloVe). At this point we have three different choices for enriching the representations using structural and sequential information: (i) structure first followed by sequence (ii) sequence first followed by structure or (iii) structure and sequence in parallel. We depict these three choices pictorially in Figure \ref{img:sss-design} and describe them below with appropriate names for future reference.

\subsubsection{Sequence contextualized GCN (\textit{Seq-GCN})}
\textit{Seq-GCN} is similar to the model proposed in \cite{bastings-etal-2017-graph,marcheggiani-titov-2017-encoding} where the word representations $x_1, x_2, x_3, \dots, x_m$ are first fed through a BiLSTM to obtain sequence contextualized representations as shown below.
\begin{equation*}
        h_{i}^{seq} = BiLSTM(h_{i-1}^{seq}, x_i)
\label{eqn:c-gcns-a}
\end{equation*}

These representations  $h_1, h_2, h_3, \dots, h_m$ are then fed to the M-GCN along with the graph $G$ to compute a $k$-hop aggregated representation as shown below:
\begin{equation*}
        h_i^{str} = M\textnormal{-}GCN(h_{i}^{seq}, G, k)    
\label{eqn:c-gcns-b}
\end{equation*}

This final representation $h_i^{final} = h_i^{str}$ for the $i$-th word thus combines semantics, sequential and structural information in that order. This is a popular way of combining GCNs with LSTMs but our experiments suggest that this does not work well for our task. We thus explore two other variants as explained below.

\subsubsection{Structure contextualized LSTM (\textit{Str-LSTM})}
Here, we first feed the word representations $x_1, x_2, x_3, \dots, x_m$ to M-GCN to obtain structure aware representations as shown below. 
\begin{equation*}
        h_i^{str} = M\textnormal{-}GCN(x_{i}, G, k)    
\label{eqn:g-lstms-a}    
\end{equation*}

These structure aware representations are then passed through a BiLSTM to capture sequence information as shown below:
\begin{align*}
        h_{i}^{seq} = BiLSTM(h_{i-1}^{seq},  h_i^{str})
\label{eqn:g-lstms-b}    
\end{align*}
This final representation $h_i^{final} = h_i^{seq}$ for the $i$-th word thus combines semantics, structural and sequential information in that order.

\subsubsection{Parallel GCN-LSTM (\textit{Par-GCN-LSTM})}
Here, both M-GCN and BiLSTMs are fed with word embeddings $x_i$ as input to aggregate structural and sequential information independently as shown below:
\begin{align*}
        h_i^{str} &= M\textnormal{-}GCN(x_{i}, G, k) \\
        h_{i}^{seq} &= BiLSTM(h_{i-1}^{seq}, x_i)
\end{align*}

The final representation, $h_i^{final}$, for each word is computed as $h_i^{final} =  h_{i}^{str}  + h_{i}^{seq}$ and combines structural and sequential information in parallel as opposed to a serial combination in the previous two variants. 

\subsection{Decoder, Attention and Copy Mechanism}
Once the final representation for each word is computed, an attention weighted aggregation, $c_t$, of these representations is fed to the decoder at each time step $t$. The decoder itself is a LSTM which computes a new state vector $s_t$ at every timestep $t$ as
\begin{align*}
s_t = LSTM (s_{t-1}, c_t)
\end{align*}

The decoder then uses this $s_t$ to compute a distribution over the vocabulary where the probability of the $i$-th word in the vocabulary is given by $p_i = softmax(Vs_t + Wc_t + b)_i$. In addition, the decoder also has a copy mechanism wherein, at every timestep $t$, it could either choose the word with the highest probability $p_i$ or copy that word from the input which was assigned the highest attention weight at timestep $t$. Such copying mechanism is useful in tasks such as ours where many words in the output are copied from the document $D$. We refer the reader to the GTTP paper for more details of the standard copy mechanism.

\section{Experimental setup}
\label{sec:experiments}
In this section, we briefly describe the dataset and task setup followed by the pre-processing steps we carried to obtain different linguistic graph structures on this dataset. We then describe the different baseline models. 

\subsection{Dataset description}
We evaluate our models using Holl-E, an English language movie conversation dataset \cite{moghe-etal-2018-towards} which contains $\sim$ 9k movie chats and $\sim$ 90k utterances. Every chat in this dataset is associated with a specific background knowledge resource from among the plot of the movie, the review of the movie, comments about the movie, and occasionally a fact table. Every even utterance in the chat is generated by copying and or modifying sentences from this unstructured background knowledge. The task here is to generate/retrieve a response using conversation history and appropriate background resource. Here, we focus only on the \textit{oracle} setup where the correct resource from which the response was created is provided explicitly. We use the same train, test, and validation splits as provided by the authors of the paper. 

\subsection{Construction of linguistic graphs}
We consider leveraging three different graph-based structures for this task. Specifically, we evaluate the popular syntactic word dependency graph (\textit{Dep-G}), entity co-reference graph (\textit{Coref-G}) and entity co-occurrence graph (\textit{Ent-G}). Unlike the word dependency graph, the two entity level graphs can capture dependencies that may span across sentences in a document. We use the dependency parser provided by SpaCy\footnote{https://spacy.io/} to obtain the dependency graph (\textit{Dep-G}) for every sentence. For the construction of the co-reference graph (\textit{Coref-G}), we use the NeuralCoref model \footnote{https://github.com/huggingface/neuralcoref. Code at https://github.com/nikitacs16/horovod\_gcn\_pointer\_generator } integrated with SpaCy. 
For the construction of the entity graph (\textit{Ent-G}), we first perform named-entity recognition using SpaCy and connect all the entities that lie in a window of $k=20$. 

\subsection{Baselines}
We categorize our baseline methods as follows:  \\
\textbf{Without Background knowledge}: We consider the simple Sequence-to-Sequence (S2S) \cite{neural_conversation_model} architecture that conditions the response generation only on the \textit{previous} utterance and completely ignores the other utterances as well as the background document. We also consider HRED \cite{HRED}, a hierarchical variant of the S2S architecture which conditions the response generation on the entire conversation history in addition to the last utterance. Of course, we do not expect these models to perform well as they completely ignore the background knowledge but we include them for the sake of completeness.\\
\textbf{With Background Knowledge}: To the S2S architecture we add an LSTM encoder to encode the document. The output is now conditioned on this representation in addition to the previous utterance. We refer to this architecture as S2S-D. Next, we use GTTP \cite{see-etal-2017-get} which is a variant of the S2S-D architecture with a copy-or-generate decoder; at every time-step, the decoder decides to copy from the background knowledge or generate from the fixed vocabulary. We also report the performance of the BiRNN + GCN architecture that uses dependency graph only as discussed in \cite{marcheggiani-titov-2017-encoding}. 
Finally, we note that in our task many words in the output need to be copied sequentially from the input background document which makes it very similar to the task of span prediction as used in Question Answering. We thus also evaluate BiDAF \cite{BIDAF}, a popular question-answering architecture, that extracts a span from the background knowledge as a response using complex attention mechanisms. For a fair comparison, we evaluate the spans retrieved by the model against the ground truth responses.

We use BLEU-4 and ROUGE (1/2/L) as the evaluation metrics as suggested in the dataset paper. Using automatic metrics is more reliable in this setting than the open domain conversational setting as the variability in responses is limited to the information in the background document. We provide implementation details in the Appendix A.

\section{Results and Discussion}

\label{sec:analysis}
In Table  \ref{table:baselines_and_best_model}, we compare our architecture against the baselines as discussed above. \textit{SSS}(BERT) is our proposed architecture in terms of the \textit{SSS} framework. We report best results within \textit{SSS} chosen across 108 configurations comprising of four different graph combinations, three different contextual and structural infusion methods, three M-GCN layers, and, three embeddings. 
\begin{table}
\begin{center}
\resizebox{\linewidth}{!}{%

\begin{tabular}{|c|c|c|c|c|}
\hline
Model   & BLEU  & \multicolumn{3}{c|}{ROUGE} \\ \hline
        &       & 1       & 2       & L      \\ \hline
S2S     & 4.63  & 26.91   & 9.34    & 21.58  \\ 
HRED    & 5.23  & 24.55   & 7.61    & 18.87  \\ \hline
S2S-D    & 11.71 & 26.36   & 13.36   & 21.96  \\ 
GTTP    & 13.97 & 36.17   & 24.84   & 31.07  \\ 
BiRNN+GCN  &  14.70 & 36.24   & 24.60    &   31.29 \\
BiDAF   & 16.79 & 26.73   & 18.82   & 23.58  \\ \hline
\textit{SSS}(GloVe) & 18.96 & 38.61   & 26.92   & 33.77  \\ 
\textit{SSS}(ELMo) & 19.32 & 39.65   & 27.37   & 34.86  \\ 
\textit{SSS}(BERT) & \textbf{22.78} & \textbf{40.09}   & \textbf{27.83}   & \textbf{35.20}   \\ \hline
\end{tabular}
}
\caption{Results of automatic evaluation. Our proposed architecture \textit{SSS}(BERT) outperforms the baseline methods.}
\label{table:baselines_and_best_model}
\end{center}
\end{table}
The best model was chosen based on performance of the validation set. From Table \ref{table:baselines_and_best_model}, it is clear that our improvements in incorporating structural and sequential information with BERT in the \textit{SSS} encoder framework significantly outperforms all other models. 

\subsection{Qualitative Evaluation}
We conducted human evaluation for the \textit{SSS} models from Table \ref{table:baselines_and_best_model} against the generated responses of GTTP. We presented 100 randomly sampled outputs to three different annotators. The annotators were asked to pick from four options: A, B, both, and none. The annotators were told these were conversations between friends. Tallying the majority vote, we obtain win/loss/both/none for \textit{SSS}(BERT) as 29/25/29/17,  \textit{SSS}(GloVe) as 24/17/47/12 and \textit{SSS}(ELMo) as 22/23/41/14. This suggests qualitative improvement using \textit{SSS} framework. We also provide some generated examples in the Appendix B1. We found that the \texttt{SSS} framework had less confusion in generating the opening responses than the GTTP baseline. These ``conversation starters'' have a unique template for every opening scenario and thus have different syntactic structures respectively. We hypothesize that the presence of dependency graphs over these respective sentences helps to alleviate the confusion as seen in Example 1. The second example illustrates why incorporating structural information is important for this task. We also observed that \textit{SSS} encoder framework does not improve on the aspects of human creativity such as diversity, initiating a context-switch, and common sense reasoning as seen in Example 3.


\begin{table}
\begin{center}
\resizebox{\linewidth}{!}{%

\begin{tabular}{|c|c|c|c|c|c|}
\hline
Emb              & Paradigm        & BLEU  & \multicolumn{3}{c|}{ROUGE} \\ \hline
                       &                 &       & 1       & 2       & L      \\ \hline
\multirow{3}{*}{GloVe} & Sem             & 4.4   & 29.72   & 11.72   & 22.99  \\  
                       & Sem+Seq         & 14.83 & 36.17   & 24.84   & 31.07  \\  
                       & \textit{SSS}& 18.96 & 38.61   & 26.92   & 33.77  \\ \hline
\multirow{3}{*}{ELMo}  & Sem             & 14.36 & 32.04   & 18.75   & 26.71  \\  
                       & Sem+Seq         & 14.61 & 35.54   & 24.58   & 30.71  \\  
                       & \textit{SSS}& 19.32 & 39.65   & 27.37   & 34.86  \\ \hline
\multirow{3}{*}{BERT}  & Sem             & 11.26 & 33.86   & 16.73   & 26.44  \\  
                       & Sem+Seq         & 18.49 & 37.85   & 25.32   & 32.58  \\  
                       & \textit{SSS}& 22.78 & 40.09   & 27.83   & 35.2   \\ \hline
\end{tabular}
}    
\end{center}

\caption{Performance of components within the \textit{SSS}
framework.}
\label{table:SSS-Framework}
\end{table}

\subsection{Ablation studies on the \textit{SSS} framework}
We report the component-wise results for the \textit{SSS} framework in Table \ref{table:SSS-Framework}. The \textit{Sem} models condition the response generation directly on the word embeddings. We observe that ELMo and BERT perform much better than GloVe embeddings. 

The \textit{Sem+Seq} models condition the decoder on the representation obtained after passing the word embeddings through the LSTM layer. These models outperform their respective \textit{Sem} models. The gain with ELMo is not significant because the underlying architecture already has two BiLSTM layers whcih are anyways being fine-tuned for the task. Hence the addition of one more LSTM layer may not contribute to learning any new sequential word information.
It is clear from Table \ref{table:SSS-Framework} that the \textit{SSS} models, that use structure information as well, obtain a significant boost in performance, validating the need for incorporating all three types of information in the architecture.


\begin{table*}
\begin{center}
\resizebox{\linewidth}{!}{%

\begin{tabular}{|c|cccc|cccc|cccc|}
\hline
Emb & \multicolumn{4}{c|}{Seq-GCN}                                                                       & \multicolumn{4}{c|}{Str-LSTM}                                                                      & \multicolumn{4}{c|}{Par-GCN-LSTM}                                                                   \\ \hline
          & \multicolumn{1}{c|}{BLEU} & \multicolumn{3}{c|}{ROUGE}                                               & \multicolumn{1}{|c|}{BLEU} & \multicolumn{3}{c|}{ROUGE}                                               & \multicolumn{1}{c|}{BLEU} & \multicolumn{3}{c|}{ROUGE}                                               \\ \hline
          & \multicolumn{1}{c|}{}     & \multicolumn{1}{c|}{1} & \multicolumn{1}{c|}{2} & \multicolumn{1}{c|}{L} & \multicolumn{1}{c|}{}     & \multicolumn{1}{c|}{1} & \multicolumn{1}{c|}{2} & \multicolumn{1}{c|}{L} & \multicolumn{1}{c|}{}     & \multicolumn{1}{c|}{1} & \multicolumn{1}{c|}{2} & \multicolumn{1}{c|}{L} \\ \hline
GloVe     & 15.61                     & 36.6                   & 24.54                  & 31.68                  & 18.96                     & 38.61                  & 26.92                  & 33.77                  & 17.1                      & 37.04                  & 25.70                  & 32.2                   \\ \hline
ELMo      & 18.44                     & 37.92                  & 26.62                  & 33.05                  & 19.32                     & 39.65                  & 27.37                  & 34.86                  & 16.35                     & 37.28                  & 25.67                  & 32.12                  \\ \hline
BERT      & 20.43                     & 40.04                  & 26.94                  & 34.85                  & 22.78                     & 40.09                  & 27.83                  & 35.20                  & 21.32                     & 39.9                   & 27.60                  & 34.87                  \\ \hline
\end{tabular}}
\end{center}
\caption{Performance of different hybrid architectures to combine structural information with sequence information}
\label{table:SSS-Models}
\end{table*}

\subsection{Combining structural and sequential information}
The response generation task of our dataset is a span based generation task where phrases of text are expected to be copied or generated as they are. The sequential information is thus crucial to reproduce these long phrases from background knowledge. This is strongly reflected in Table \ref{table:SSS-Models} where \textit{Str-LSTM} which has the LSTM layer on top of GCN layers performs the best across the hybrid architectures discussed in Figure \ref{img:sss-design}. The \textit{Str-LSTM} model can better capture sequential information with structurally and syntactically rich representations obtained through the initial GCN layer. The \textit{Par-GCN-LSTM} model performs second best. However, in the parallel model, the LSTM cannot leverage the structural information directly and relies only on the word embeddings. \textit{Seq-GCN} model performs the worst among all the three as the GCN layer at the top is likely to modify the sequence information from the LSTMs.


\subsection{Understanding the effect of structural priors}

While a combination of intra-sentence and inter-sentence graphs is helpful across all the models, the best performing model with BERT embeddings relies only on the dependency graph. In case of GloVe based experiments, the entity and co-reference relations were not independently useful with the \textit{Str-LSTM} and \textit{Par-GCN-LSTM} models, but when used together gave a significant performance boost, especially for \textit{Str-LSTM}. However, most of the BERT based and ELMo based models achieved competitive performance with individual entity and co-reference graphs. There is no clear trend across the models. Hence, probing these embedding models is essential to identify which structural information is captured implicitly by the embeddings and which structural information needs to be added explicitly. For the quantitative results, please refer to the Appendix B2.

\subsection{Structural information in deep contextualised representations}

Earlier work has suggested that deep contextualized representations capture syntax and co-reference relations \cite{peters-etal-2018-dissecting,jawahar-etal-2019-bert,tenney-etal-2019-bert, hewitt-manning-2019-structural}.
We revisit Table \ref{table:SSS-Framework} and consider the \textit{Sem+Seq} models with ELMo and BERT embeddings as two architectures that \textit{implicitly} capture structural information. 
We observe that the \textit{SSS} model using the simpler GloVe embedding outperforms the ELMo \textit{Sem+Seq} model and performs slightly better than the BERT \textit{Sem+Seq} model. 

Given that the \textit{SSS} models outperform the corresponding \textit{Sem+Seq} model, the extent to which the deep contextualized word representations learn the syntax and other linguistic properties implicitly is questionable. Also, this calls for better loss functions for learning deep contextualised representations that can incorporate structural information explicitly.

More importantly, all the configurations of \textit{SSS} (GloVe) have lesser memory footprint in comparison to both ELMo and BERT based models. Validation and training of GloVe models require one-half, sometimes even one-fourth of computing resources.  Thus, the simple addition of structural information through the GCN layer to the established Sequence-to-Sequence framework that can perform comparably to stand-alone expensive models is an important step towards Green AI\cite{green_ai}.

\section{Conclusion}
We demonstrated the usefulness of incorporating structural information for the task of background aware dialogue response generation. We infused the structural information explicitly in the standard semantic+sequential model and observed performance boost. We studied different structural linguistic priors and different ways to combine sequential and structural information.  We also observe that explicit incorporation of structural information helps the richer deep contextualized representation based architectures. We believe that the analysis presented in this work would serve as a blueprint for analysing future work on GCNs ensuring that the gains reported are robust and evaluated across different configurations.

\appendix
\section{Implementation Details}
\subsection{Base Model}
The baseline in \cite{moghe-etal-2018-towards} adapted the architecture of Get to the Point \cite{see-etal-2017-get} for background aware dialogue response generation task. 
In the summarization task, the input is a \textit{document} and the output is a \textit{summary} whereas in our case the input is a \{\textit{resource/document, context}\} pair and the  output is a \textit{response}. Note that the context includes the previous two utterances (dialog history) and the current utterance. Since, in both the tasks, the output is a sequence (\textit{summary} v/s \textit{response}) we don't need to change the decoder (\textit{i.e.}, we can use the decoder from the original model as it is). However, we need to change the input fed to the decoder. 
We use an RNN to compute a representation of the conversation history. Specifically, we consider the previous $k$ utterances as a single sequence of words and feed these to an RNN. Let $M$ be the total length of the context (\textit{i.e.}, all the $k$ utterances taken together) then the RNN computes representations $h_1^d, h_2^d, ..., h_M^d$ for all the words in the context. The final representation of the context is then the attention weighted sum of these word representations:
\begin{equation}
\begin{split}
f_i^t &= v^T tanh(W_c h_i^d + V s_t + b_d)\\
m^t &= softmax(f^t)\\
d_t &= \sum_i m_i^t h_i^d
\end{split}
\end{equation}
Similar to the original model, we use an RNN to compute the representation of the document. Let $N$ be the length of the document then the RNN computes representations $h_1^r, h_2^r, ..., h_N^r$ for all the words in the resource (we use the superscript $r$ to indicate resource). We then compute the query aware resource representation as follows. 
\begin{equation}
\label{new_attention_on_document}
\begin{split}
e_i^t &= v^T tanh(W_r h_i^r + U s_t + V d_t+ b_r)\\
a^t &= softmax(e^t)\\
c_t &= \sum_i a_i^t h_i^r
\end{split}
\end{equation}
where $c_t$ is the attended context representation. Thus, at every decoder time-step, the attention on the document words is also based on the currently attended context representation. 

The decoder then uses $r_t$ (document representation) and $s_t$ (decoder's internal state) to compute a probability distribution over the vocabulary $P_{vocab}$. In addition, the model also computes $p_{gen}$ which indicates that there is a probability $p_{gen}$ that the next word will be \textit{generated} and a probability $(1 - p_{gen})$ that the next word will be \textit{copied}. We use the following modified equation to compute $p_{gen}$
\begin{equation}
p_{gen} = \sigma(w_{r}^T r_t + w_s^T s_t + w_x^T x_t + b_{g})
\end{equation}

where $x_t$ is the previous word predicted by the decoder and fed as input to the decoder at the current time step. Similarly, $s_t$ is the current state of the decoder computed using this input $x_t$. The final probability of a word $w$ is then computed using a combination of two distributions, \textit{viz.}, ($P_{vocab}$) as described above and the attention weights assigned to the document words as shown below
\begin{equation}
P(w) = p_{gen} P_{vocab}(w) + (1 - p_{gen}) \sum_{i:w_i=w} a_i^t
\end{equation}
where $a_i^t$ are the attention weights assigned to every word in the document as computed in Equation \ref{new_attention_on_document}. Thus, effectively, the model could learn to copy a word $i$ if $p_{gen}$ is low and $a_i^t$ is high. 
This is the baseline with respect to the LSTM architecture (Sem + Seq). For, GCN based encoders, the $h_i^r$ is the final outcome after the desired GCN/LSTM configuration. 

\subsection{Hyperparameters}
We selected  the  hyper-parameters using the validation set. We used Adam optimizer with a learning rate of 0.0004 and a batch size of 64. We used GloVe embeddings of size 100. For the RNN-based encoders and decoders, we used LSTMs with a hidden state of size 256. We used gradient clipping with a maximum gradient norm of 2. We used a hidden state of size 512 for \textit{Seq-GCN} and 128 for the remaining GCN-based encoders. We ran all the experiments for 15 epochs and we used the checkpoint with the least validation loss for testing. For models using ELMo embeddings, a learning rate of 0.004 was most effective. For the BERT-based models, a learning rate of 0.0004 was suitable. Rest of the hyper-parameters  and other setup details remain the same for experiments with BERT and ELMo. Our work follows a task specific architecture as described in the previous section. Following the definitions in \cite{peters-etal-2019-tune}, we use the ``feature extraction'' setup for both ELMo and BERT based models. 

\section{Extended Results}
\subsection{Qualitative examples}
We illustrate different scenarios from the dataset to identify the strengths and weaknesses of our models under the \textit{SSS} framework in Table \ref{tab: qualitative_evaluation_SSS}. We compare the outputs from the best performing model on the three different embeddings and use GTTP as our baseline. The best performing combination of sequential and structural information for all the three models in the SSS framework is \textit{Str-LSTM}. The best performing \textit{SSS}(GloVe) and \textit{SSS}(ELMo) architectures use all the three graphs while \textit{SSS}(BERT) uses only the dependency graph. 

We find that the \textit{SSS} framework improves over the baseline for the cases of opening statements (see Example 1). The baseline had confusion in picking opening statements and often mixed the responses for ``Which is your favorite character?", ``Which is your favorite scene'' and ``What do you think about the movie?''.  The responses to these questions have different syntactic structures - ``My favorite character is XYZ'', ``I liked the one in which XYZ'', and `` I think this movie is XYZ'' where XYZ was the respective crowdsourced phrase. The presence of dependency graphs over the respective sentences may help to alleviate the confusion. 

Now consider the example under \textit{Hannibal} in Table \ref{tab: qualitative_evaluation_SSS}. We find that the presence of a co-reference graph between ``Anthony Hopkins'' in the first sentence and ``he'' in the second sentence can help in continuing the conversation on the actor ``Anthony Hopkins''. Moreover, connecting tokens in ``Anthony Hopkins'' to refer to ``he'' in the second sentence is possible because of the explicit entity-entity connection between the two tokens. However, this is applicable only to \textit{SSS}(GloVe) and \textit{SSS}(ELMo) as their best performing versions use these graphs along with the dependency graph while the best performing \textit{SSS}(BERT) only uses dependency graph and may have learnt the inter-sentence relations implicitly. 

There is a limited diversity of responses generated by the \textit{SSS} framework as it often resorts to the patterns seen during training while it is not copying from the background knowledge. We also identify that \textit{SSS} framework cannot handle the cases where Speaker2 initiates a context switch, \textit{i,e;} when Speaker2 introduces a topic that has not been discussed in the conversation so far. In the chat on \textit{The Road Warrior} in Table \ref{tab: qualitative_evaluation_SSS}, we find that \textit{Mad Max: Fury Road} has been used to initiate a discussion that compares the themes of both the movies. All the models produce irrelevant responses.

\begin{table*}[h]
\centering
\begin{tabular}{|l|p{3.9cm}|p{3.9cm}|p{3.9cm}|}
\hline
Movie        & Rocky V  (Example 1)                                                                                                                                                                            & Hannibal (Example 2)                                                                                                                                             & The Road Warrior (Example 3)                                                                                                                                                                                                         \\ \hline
Resource     & Rocky V, terribly under-rated.  I liked the one in which Rocky .... .  My favorite character was Adrian because she was  ...I think it was enjoyable, though slightly less than th. & Anthony Hopkins gave .... However, the material  he was given to work ....  almost seemed like an excuse to present us                               & Box Office: \$ 9,003,011 Similar Movies: Lethal Weapon, Mad Max: Fury Road ...                                                                                                                                           \\ \hline
Chat         & Speaker 1: Which is your favourite character in this?                                                                                                                               & Speaker 1: What is your opinion about the movie? \newline Speaker 2: I think it was a fair sequel. \newline Speaker 1: Anthony Hopkins gave an impeccable performance. & Speaker 1: Which is your favourite character in this? \newline Speaker 2: My favorite character was Dog \newline Speaker 1: I liked the feral kid even after he's found in the car and Max evicts him without ceremony he doesn't give up. \\ \hline
Ground Truth & my favorite character was adrian because she was strong and did her best to keep her family together through tough times .                                                          & however , the material he was given to work with was not as good as silence of the lambs .                                                           & Have you seen the new Mad Max: Fury Road?                                                                                                                                                                                \\ \hline
Baseline     & i liked the one in which rocky and gunn fight in the street .                                                                                                                       & in fairness , perhaps there was no way it could be .                                                                                                 & lethal weapon .                                                                                                                                                                                                          \\ \hline
\textit{SSS}(GloVe)   & my favorite character was adrian because she was strong and did her best to keep her together through tough times .                                                                 & however , the material he was given to work with was not as good as silence of the lambs .                                                           & it made \$ 9,003,011                                                                                                                                                                                                     \\ \hline
\textit{SSS}(ELMo)    & my favorite character was adrian because she was strong and did her best to keep her family together through tough times .                                                          & the material he was given to work with was not as good as silence of the lambs .                                                                     & {[}UNK{]} only one man can make the difference in the future .                                                                                                                                                           \\ \hline
\textit{SSS}(BERT)    & my favorite character was adrian because she was strong and did her best to keep her family together through tough times .                                                          & the material he was given to work with was not as good as silence of the lambs .                                                                     & yes .{[}UNK{]} only one man can make the difference in the future .                                                                                                                                                      \\ \hline

\end{tabular}
\caption{Sample outputs from the SSS framework compared with baseline and ground truth responses.}
\label{tab: qualitative_evaluation_SSS}
\end{table*}

\subsection{Quantitative results}
We explore the effect of using different graphs in Table \ref{table:SSS-Modelss}.
\begin{table*}[t]
\centering
\resizebox{\linewidth}{!}{%

\begin{tabular}{|c|c|c|c|c|c|c|c|c|c|c|c|c|}
\hline
Graph                                                    & \multicolumn{4}{c|}{GloVe}         & \multicolumn{4}{c|}{ELMo}          & \multicolumn{4}{c|}{BERT}          \\ \hline
                                                         & BLEU  & \multicolumn{3}{c|}{ROUGE} & BLEU  & \multicolumn{3}{c|}{ROUGE} & BLEU  & \multicolumn{3}{c|}{ROUGE} \\ \hline
                                                         &       & 1       & 2       & L      &       & 1       & 2       & L      &       & 1       & 2       & L      \\ \hline
Dep                                                      & 16.79 & 37.77   & 25.89   & 32.88  & 17.00 & 37.56   & 26.14   & 32.77  & 22.78 & 40.09   & 27.83   & 35.2   \\ \hline
Dep+Ent                                                  & 14.44 & 35.14   & 24.61   & 30.43  & 18.34 & 39.55   & 28.00   & 34.76  & 19.33 & 39.37   & 27.52   & 34.33  \\ \hline
Dep+Coref                                                & 16.58 & 37.60   & 25.72   & 32.63  & 18.56 & 40.08   & 28.42   & 35.06  & 20.99 & 40.10   & 28.66   & 35.11  \\ \hline
\begin{tabular}[c]{@{}c@{}}Dep+Ent\\ +Coref\end{tabular} & 18.96 & 38.61   & 26.92   & 33.77  & 19.32 & 39.65   & 27.37   & 34.86  & 20.37 & 39.11   & 27.2    & 34.19  \\ \hline
\end{tabular}}
\caption{Comparing performance of different structural priors across different semantic information on the \textit{Str-LSTM} architecture.}
\label{table:SSS-Modelss}
\end{table*}

\cleardoublepage
\newpage
\bibliographystyle{unsrt}
\bibliography{emnlp2018}

\end{document}